%% file: main.tex
\documentclass[10pt,twocolumn,letterpaper]{article}

\usepackage{cvpr}              % To produce the CAMERA-READY version
\input{preamble}

\definecolor{cvprblue}{rgb}{0.21,0.49,0.74}
\usepackage[pagebackref,breaklinks,colorlinks,allcolors=cvprblue]{hyperref}
\usepackage{comment}
\usepackage{algorithm}
\usepackage{algorithmic}
\usepackage{tabularx}
\usepackage{subcaption}
\usepackage{mdframed}
\usepackage{xcolor}
\usepackage[title]{appendix}
\usepackage[accsupp]{axessibility}

\def\paperID{10246} % *** Enter the Paper ID here
\def\confName{CVPR}
\def\confYear{2025}

\title{TIDE: Training Locally Interpretable Domain Generalization Models \\ Enables Test-time Correction}

\author{
    \textit{Aishwarya Agarwal}$^{1,2}$\thanks{\scriptsize \textit{aishwarya.agarwal@research.iiit.ac.in}, \textit{aishagar@adobe.com}} \quad
    \textit{Srikrishna Karanam}$^{2}$\thanks{\scriptsize \textit{skaranam@adobe.com}} \quad
    \textit{Vineet Gandhi}$^{1}$\thanks{\scriptsize \textit{vgandhi@iiit.ac.in}}\\[2mm]
    \small $^{1}$CVIT, Kohli Centre for Intelligent Systems, IIIT Hyderabad, India\\
    \small $^{2}$Adobe Research, Bengaluru, India
}

\begin{document}
\maketitle
\input{sec/0_abstract}    
\input{sec/1_intro}
\input{sec/2_related_works}

\input{sec/3_methodology}
\input{sec/4_exps}

\input{sec/5_conclusion}

{
    \small
    \bibliographystyle{ieeenat_fullname}
    \bibliography{longstrings, main}
}

\clearpage

\input{sec/supplementary}

% \clearpage
% \input{sec/supplementary}

\end{document}

%% file: preamble.tex
\usepackage{lineno}

\newcommand{\modelname}{TIDE}

\newcommand{\tabpubb}[1]{{\tiny \color{darkgray}{#1}}}

%% file: sec/0_abstract.tex
\begin{abstract}

We consider the problem of single-source domain generalization. Existing methods typically rely on extensive augmentations to synthetically cover diverse domains during training. However, they struggle with semantic shifts (e.g., background and viewpoint changes), as they often learn global features instead of local concepts that tend to be domain invariant. To address this gap, we propose an approach that compels models to leverage such local concepts during prediction. Given no suitable dataset with per-class concepts and localization maps exists, we first develop a novel pipeline to generate annotations by exploiting the rich features of diffusion and large-language models. Our next innovation is TIDE, a novel training scheme with a concept saliency alignment loss that ensures model focus on the right per-concept regions and a local concept contrastive loss that promotes learning domain-invariant concept representations. This not only gives a robust model but also can be visually interpreted using the predicted concept saliency maps. Given these maps at test time, our final contribution is a new correction algorithm that uses the corresponding local concept representations to iteratively refine the prediction until it aligns with prototypical concept representations that we store at the end of model training. We evaluate our approach extensively on four standard DG benchmark datasets and substantially outperform the current state-of-the-art ($12\%$ improvement on average) while also demonstrating that our predictions can be visually interpreted.

\end{abstract}

%% file: sec/1_intro.tex
\section{Introduction}
\label{sec:intro}

Enhancing deep neural networks to generalize to out-of-distribution samples remains a core challenge in machine learning and computer vision research, as real-world test data often diverges significantly from the training distribution~\cite{liu2021towards}. The challenges compound when obtaining labeled samples from the target domain is expensive or unfeasible, hindering application of semi-supervised learning or domain adaptation~\cite{kouw2019review,tzeng2017adversarial,long2015learning}. The problem of Domain Generalization (DG)~\cite{wang2022generalizing,zhang2022towards,zhou2021domain,li2018deep,sivaprasad2022class} represents a promising avenue for developing techniques that capture domain-invariant patterns and improve performance on out-of-distribution samples. In this paper, we focus on Single Source Domain Generalization (SSDG), where a model is trained on data from a single domain and aims to generalize well to unseen domains~\cite{qiao2020learning,volpi2018generalizing}. It represents the most strict form of DG, as the model must extract domain-invariant features from a single, often limited perspective, without exposure to the variation present across multiple domains.

\begin{figure}[t]
    \centering
    \includegraphics[width=1.02\linewidth]{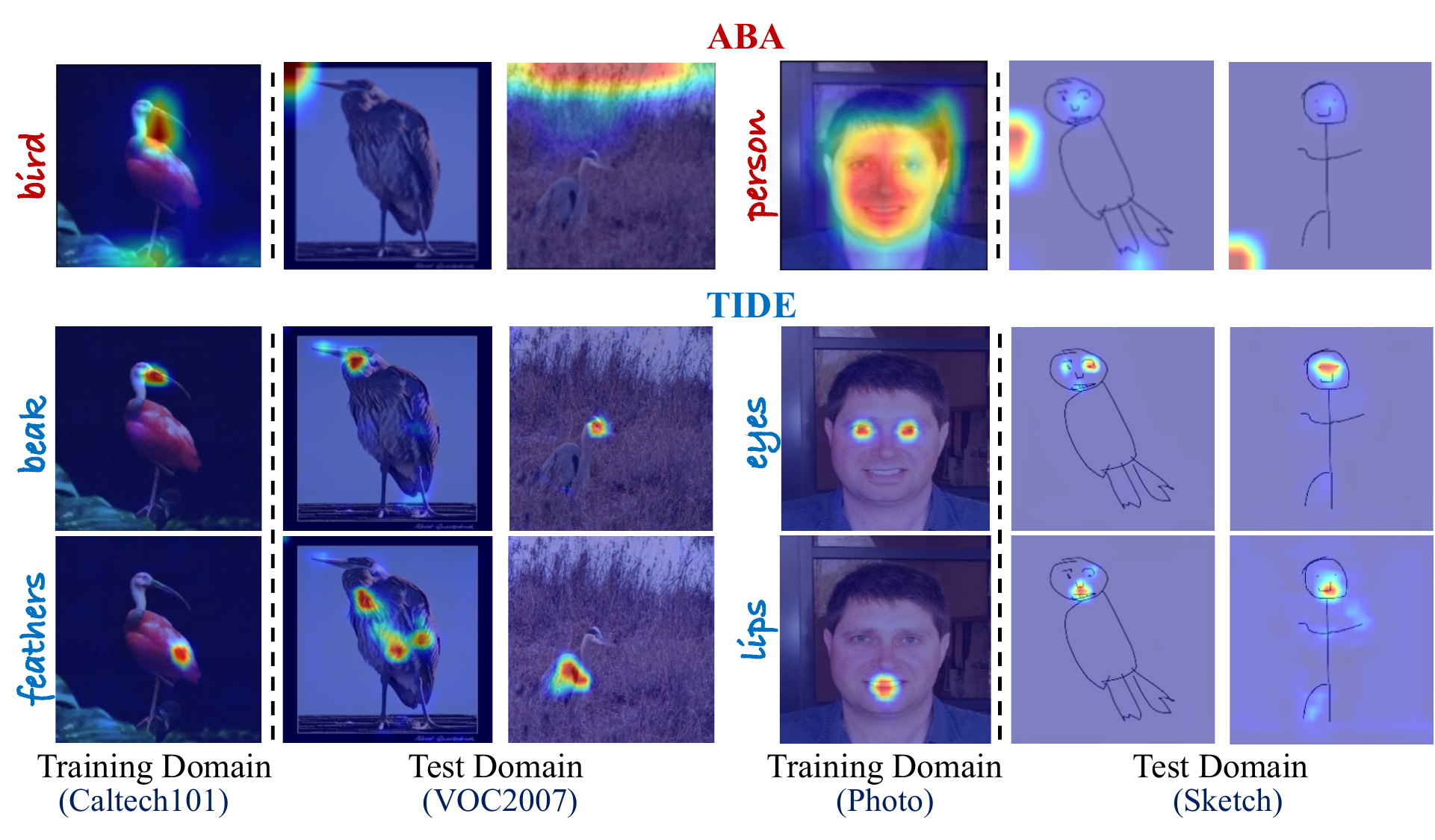}
    
 \caption{Samples from VLCS (left) and PACS dataset (right) across domain shifts, corresponding to bird and person class. First row displays GradCAM maps~\cite{selvaraju2017grad} for ABA's class predictions. We observe that model attention of ABA~\cite{cheng2023adversarial} falters across domain shifts. Second and third row display the concept specific GradCAM maps from TIDE. We posit that accurate concept learning and localization facilitates DG.} 
    \label{fig:motivation-dg-exp}
\end{figure}

Most previous work on SSDG~\cite{cheng2023adversarial,cubuk2020randaugment,hendrycks2022pixmix,yuan2022not,zhou2021domain,hendrycks2019augmix,cugu2022attention,huang2020self} relies extensively on data augmentations to support the learning of domain-generalized features. The premise is that constructing an extensive repertoire of augmentations synthesizes instances encompassing a wide spectrum of human-recognizable domains. However, accounting for all conceivable real-world augmentations presents an immense challenge. These models have shown reasonable success in addressing variations in style and texture~\cite{somavarapu2020frustratingly,zhou2024mixstyle}; however, their performance remains modest when faced with more semantic domain shifts, such as changes in background and viewpoint e.g., in the VLCS dataset~\cite{qu2023modality,torralba2011unbiased}. 

Consider the results of the state-of-the-art ABA~\cite{cheng2023adversarial} model on the two examples in Figure~\ref{fig:motivation-dg-exp}. In the first example (left), domain shifts manifest as variations in background and viewpoint. In the second example (right), the training data comprises photos of human faces, whereas the test data consists of skeletal sketches. In both cases, the domain shifts extend beyond style or texture variations, with ABA failing to correctly classify the samples. Class-level saliency maps~\cite{selvaraju2017grad} reveal that this misclassification stems from inadequate focus on \emph{critical local concepts} under domain shifts, such as \texttt{beak} and \texttt{feathers} for birds, or \texttt{eyes} and \texttt{lips} for persons. Such local concepts are integral to class definition and remain invariant across domains, and a robust DG model must adeptly capture these stable features. We posit that prior efforts learn global features per class;
and if the model fails to learn the correct feature set in the training domain, its generalization performance is compromised as noted from Figure~\ref{fig:motivation-dg-exp} above. The inconsistency in concept localization further exacerbates the interpretability and explainability of these models.

We adopt an alternative approach, wherein, rather than attempting to encompass all potential augmentations, we compel the model to leverage \emph{essential class-specific concepts} for classification. Our key idea is to force the model to attend to these concepts during training. The primary hurdle in this path is lack of annotated data specifying relevant class-specific concepts along with their corresponding localizations. To this end, our first contribution is a novel pipeline that harnesses large language models (LLMs) and diffusion models (DMs) to identify key concepts and generate concept-level saliency maps in a \emph{scalable}, \emph{automated fashion}. We demonstrate that DMs, via cross-attention layers, can generate concept-level saliency maps for generated images with notable granularity. Given these maps for a single synthesized image within a class, we leverage the rich feature space of DMs to transfer them to images across diverse domains in DG benchmarks~\cite{tang2023emergent}.

We subsequently introduce \modelname, our second contribution which employs cross-entropy losses over class and concept labels along with a novel pair of loss functions: a \emph{concept saliency alignment loss} that ensures the model attends to the correct regions for each local concept, and a \emph{local concept contrastive loss} that promotes learning domain invariant features for these regions.
As shown in Figure~\ref{fig:motivation-dg-exp}, \modelname~consistently attends to local concepts such as \texttt{beak} and \texttt{eyes} in the training domain as well as across substantial domain shifts. This precise localization significantly bolsters SSDG performance while enabling the generation of visual explanations for model decisions.

Marking a major leap in model interpretability and performance gains, our third contribution is to demonstrate how our model's concept-level localization can be effectively leveraged for \emph{performance verification} and \emph{test-time correction}. To this end, we introduce the notion of \emph{local concept signatures}: prototypical features derived by pooling concept-level features across training samples, guided by corresponding saliency maps. If the concept features associated with the predicted class label do not align with their signatures, it signals the use of incorrect features for prediction. Consequently, we employ iterative refinement through concept saliency masking until concept predictions align with their corresponding signatures. More formally, our paper makes following contributions: 

\begin{itemize}
    \item We propose a novel synthetic annotation pipeline to automatically identify key per-class local concepts and their localization maps, and transfer them to real images found in DG benchmarks.
    \item We propose a novel approach, TIDE, to train locally interpretable models with a novel concept saliency alignment loss that ensures accurate concept localization and a novel local concept contrastive loss that ensures domain-invariant features. 
    \item We show TIDE enables correcting model prediction at test time using the predicted concept saliency maps as part of a novel iterative attention refinement strategy.

    \item We report extensive results on four standard DG benchmarks, outperforming the current state-of-the-art by a significant $12\%$ on average. 
     
\end{itemize}

%% file: sec/2_related_works.tex
\section{Related Works}
\label{sec:relatedWorks}

Multi Source domain generalization (MSDG) methods assume access to multiple source domains and domain-specific knowledge during training and have proposed ways to learn multi-source domain-invariant features ~\cite{du2020learning,wang2019learning,motiian2017unified,ghifary2015domain,muandet2013domain}, utilize domain labels to capture domain shifts~\cite{xiao2021bit,d2019domain,muandet2013domain}, and design target-specific augmentation strategies~\cite{gokhale2021attribute,geirhos2018imagenet,jaderberg2015spatial}. However, these assumptions are not practical for real-world applications~\cite{xu2023simde,sivaprasad2022class} and we instead focus on the more challenging SSDG setting, where only one source domain and no prior target knowledge is available.

\begin{figure*}[t]
    \centering
    \includegraphics[width = 1.01\linewidth]{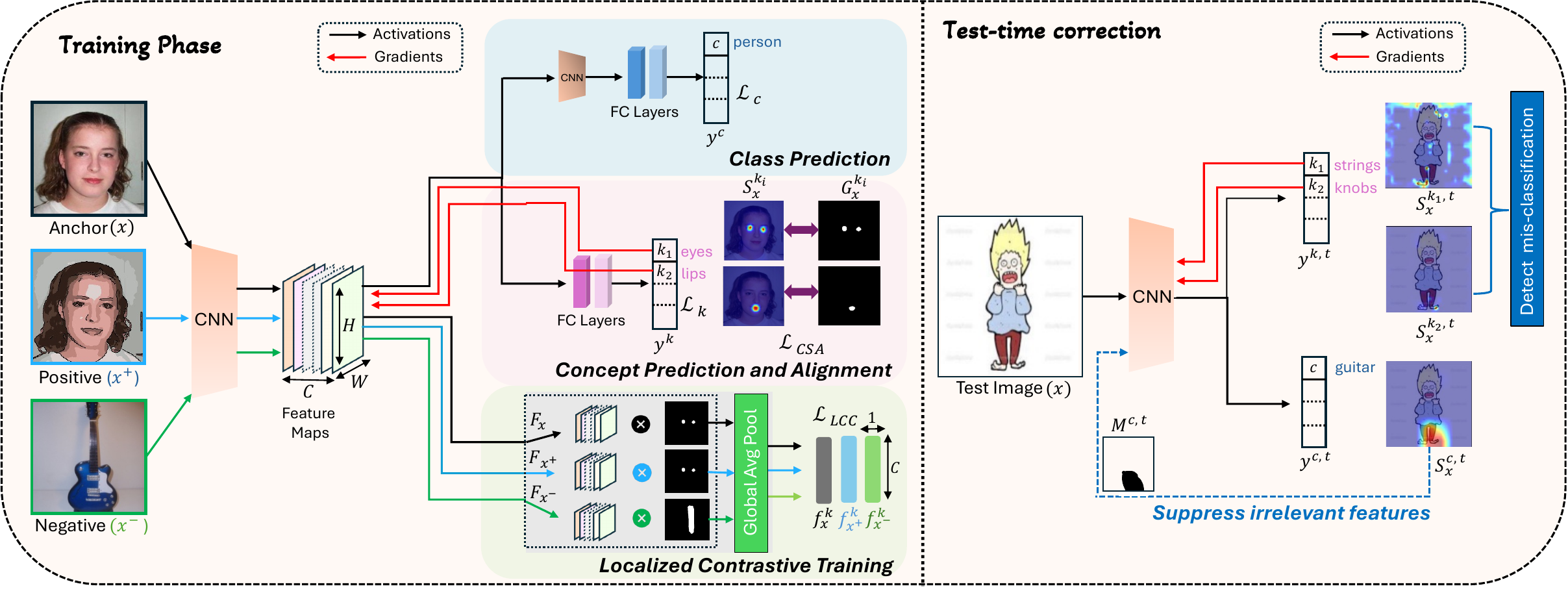}
    \caption{The TIDE pipeline: Left—Training on a single domain with cross-entropy losses for class ($\mathcal{L}_{\text{c}}$) and concept labels ($\mathcal{L}_{\text{k}}$), alongside Concept Saliency Alignment ($\mathcal{L}_{\text{CSA}}$) and Local Concept Contrastive losses ($\mathcal{L}_{\text{LCC}}$). Right—Test-time correction strategy applied in TIDE.  }
    \label{fig:pipelineFig}
\end{figure*}

Most SSDG approaches have used augmentations to improve DG \cite{cheng2023adversarial,cubuk2020randaugment,hendrycks2022pixmix,yuan2022not,zhou2021domain,hendrycks2019augmix,cugu2022attention,huang2020self}. While \cite{yuan2022not} also uses diffusion models (DMs)~\cite{rombach2022high} for data augmentation, our approach differs by leveraging the rich feature space of DMs primarily for offline saliency map annotation instead of augmentation. More recent work in SSDG \cite{li2024prompt,chen2023meta,li2022uncertainty} has also turned to approaches based on domain-specific prompting and causal inference instead of data augmentations. However, these methods learn global features, limiting invariance across unseen domains. Finally, these methods also do not provide any signals to interpret model predictions.

On the other hand, while methods like Concept Bottleneck Models (CBMs) \cite{koh2020concept,margeloiu2021concept} also learn local concepts, they are restricted to a predefined set of concepts and more crucially, cannot ground them to image regions that these concepts represent. Recent methods \cite{yang2023language,oikarinen2023label} have scaled CBMs to a large number of classes and also proposed to ways to learn multi-scale representations \cite{wang2024mcpnet}, but they are also unable to connect these concepts to image pixels. Our SSDG method addresses these challenges by producing local concept saliency maps, enhancing both DG performance and model prediction interpretability.

%% file: sec/3_methodology.tex
\section{Methodology}
\label{sec:method}

The proposed framework is depicted in Figure~\ref{fig:pipelineFig}. Training comprises three components: class prediction, concept prediction with saliency alignment, and localized contrastive training. During inference, class and concept predictions are integrated with a test-time correction mechanism.

In the concept alignment phase, we employ concept-level saliency maps as ground truth to enforce focus on interpretable features, using a saliency alignment loss. Domain invariance is further promoted by localized contrastive training, where we train the model to 
cluster similar concepts (e.g., \texttt{eyes} across augmentations) while separating unrelated concepts (e.g., \texttt{strings}). Finally, a test-time correction mechanism iteratively refines attention by suppressing irrelevant regions, leveraging concept-signatures to redirect focus and improve classification accuracy. We proceed with a concise review of the key notations, followed by an in-depth exposition of our approach and the pipeline for generating ground truth concept-level annotations.

We assume the source data is observed from a single domain, denoted as $D = \{{x_i, y^c_i, y^k_i}\}_{i=1}^N$, where $x_i$ represents the $i$-th image, $y^c_i$ the class label, $y^k_i$ the concept label, and $N$ the total sample count in the source domain. The shared CNN backbone is used to obtain the backbone features $F_{x} \in \mathbb{R}^{W\times H\times C}$. The automatically generated ground truth concept-level saliency maps are denoted by $G^k_{x}$. The GradCAM~\cite{selvaraju2017grad} saliency maps corresponding to class and concepts labels are denoted by $S^c_x$ and $S^k_x$ respectively.

\subsection{Generating Concept-level Annotations}
\label{subsec:genAnnots}

During training, we aim our model to identify and spatially attend to stable, discriminative regions. However, existing DG datasets lack fine-grained, concept-level annotations for such regions. To address this, we propose a novel pipeline that uses LLMs and DMs to automate scalable concept-level saliency map generation.

\begin{figure}[t]
    \centering
    \includegraphics[width = 0.95\linewidth]{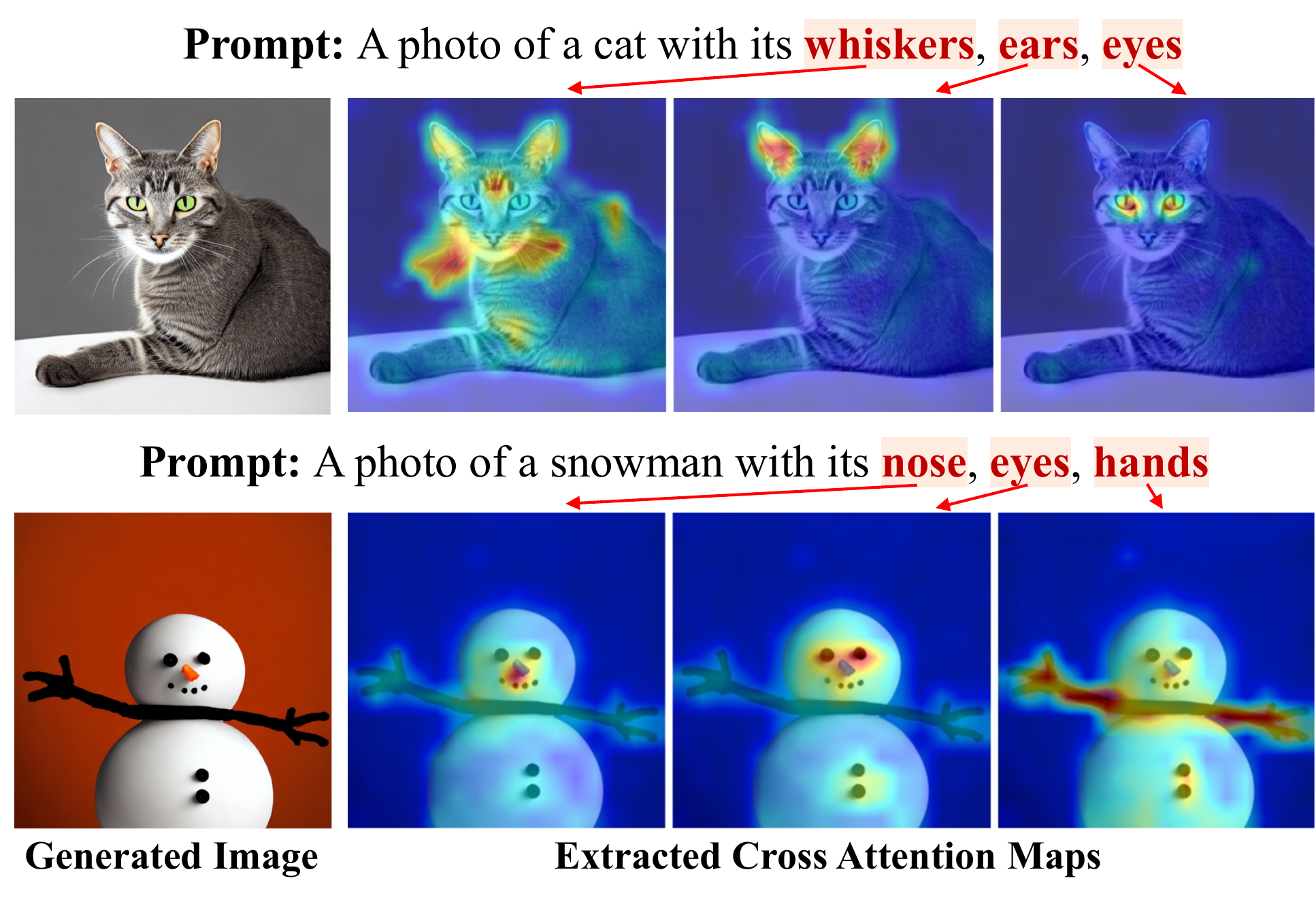}
    \caption{ The first column displays the image generated from the given prompt, while the subsequent three columns show the cross-attention maps corresponding to each concept in the prompt. }
    \label{fig:caMapsSyn}
\end{figure}

Our primary insight is that DMs can be harnessed to generate high-quality, concept-level saliency maps for synthesized images. Extracting cross-attention maps~\cite{agarwal2023star,chefer2023attend} from a DM, given the prompt with concepts and corresponding synthesized image, yields highly granular concept-level saliency maps. Figure~\ref{fig:caMapsSyn} demonstrates how DMs emphasize specific regions for distinct concepts, capturing fine-grained attention to features such as a cat's \texttt{whiskers} or a snowman's \texttt{hands}. With this technique serving as an efficient tool for yielding concept maps for synthesized images, the ensuing questions are: (i) how to automate the identification of pertinent concepts for each class across datasets, and (ii) how to transfer these concept maps to real-world images to generate ground-truth annotations.

First, to identify the key concepts associated with each class, we use GPT-3.5~\cite{brown2020language}, which we prompt to generate a list of distinctive, stable features (prompt in the supplement) for each class. For instance, GPT-3.5 outputs concepts such as \texttt{whiskers}, \texttt{ears}, and \texttt{eyes} for a cat. We generate a prompt leveraging these concepts, which is then used to synthesize an exemplar image for each class. We further derive the corresponding concept-level attention maps as outlined in the preceding paragraph.

\begin{figure}[t]
    \centering
    \includegraphics[width = 0.95\linewidth]{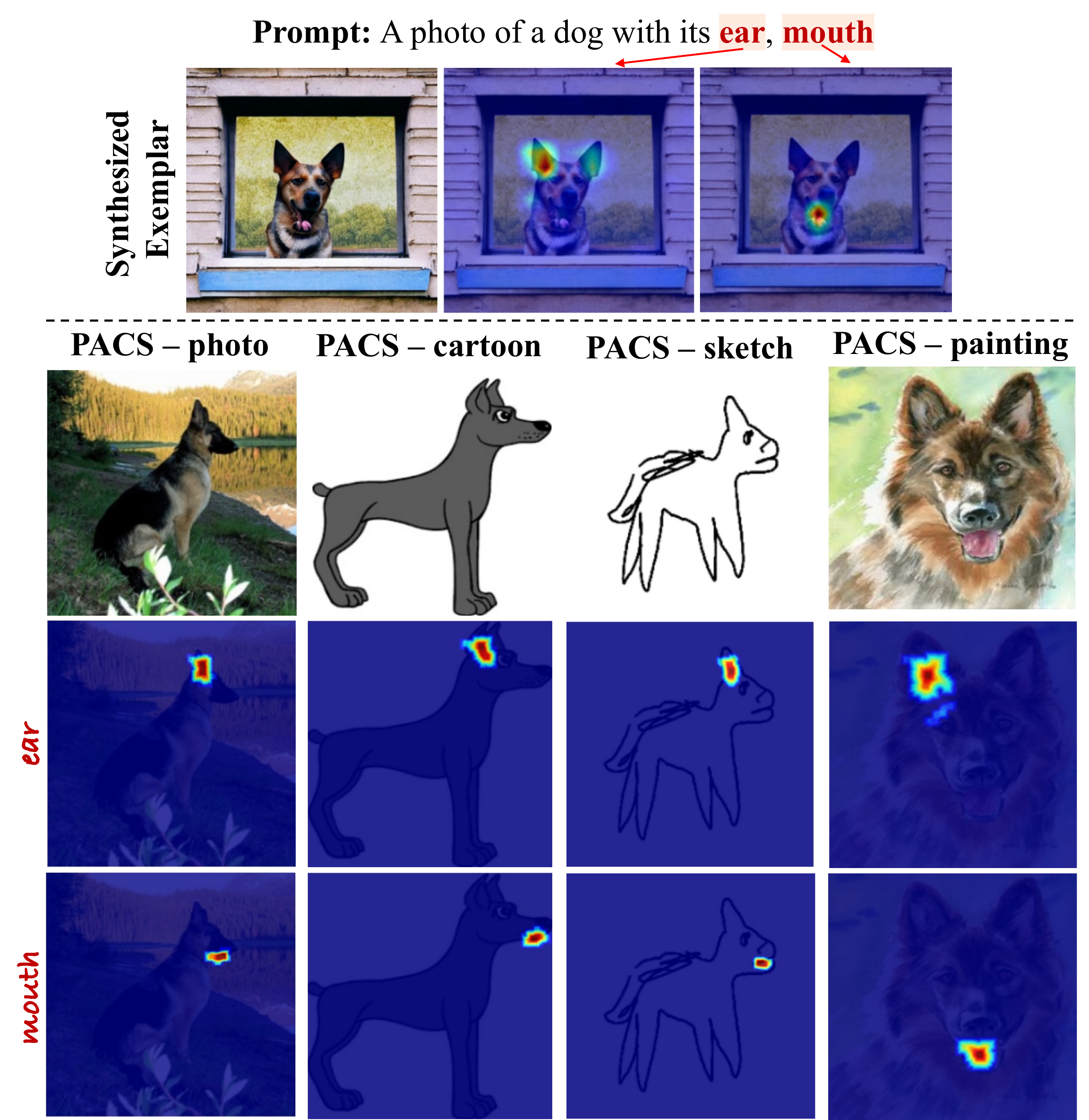}
    \caption{The first row presents the prompt, corresponding synthesized image and attention maps for the concepts \texttt{ear} and \texttt{mouth}. Below, we demonstrate that using diffusion features correspondences these concept saliency maps from a single exemplar can be automatically transferred on dog images across domains.}
    \label{fig:diftPACS}
\end{figure}

Given a single synthesized exemplar image for each class and the concept-level saliency maps, we turn to the task of transferring these saliency annotations to real-world images. DM’s feature space is particularly well-suited for this, as it captures detailed, semantically-rich representations that allow us to match synthetic concept regions to real images across domains. Leveraging this, we use the Diffusion Feature Transfer (DIFT) method~\cite{tang2023emergent}, which computes pixel-level correspondences by comparing cosine similarities between DM features. Through this approach, we establish a region-to-region correspondence between synthetic saliency maps (e.g., \texttt{mouth} of a dog) and similar regions in real-world images. This enables us to generate comprehensive, consistent concept-level annotations across domains as shown in Figure~\ref{fig:diftPACS}.

We obtain these concept-level annotations for widely-used benchmarks, including VLCS~\cite{fang2013unbiased}, PACS~\cite{li2017deeper}, DomainNet~\cite{peng2019moment}, and Office-Home~\cite{venkateswara2017deep}. For an image $x$, the binarized concept-level maps ($G^k_x$) are henceforth referred to as ground-truth saliency maps (GT-maps) in this paper. Having established a process for generating saliency maps, the next challenge is identifying the subset of concepts that the model actually relies on for making predictions.

\subsubsection{Concepts that matter}

We restrict our method to essential concepts, filtering out those that do not contribute meaningfully to classification. To do this, we train a ResNet-18~\cite{he2016deep} classifier on the source domain~\cite{krizhevsky2012imagenet}, compute GradCAM~\cite{selvaraju2017grad} saliency maps for each class label, and use them to identify regions the model focuses on when making predictions. 

We compute the overlap between the GradCAM maps and GT-maps for each known concept in the dataset. Given image $x$, its saliency map $S^c_x$ for class $c$ and the GT-map $G^k_x$ for concept $k$, we define the overlap $O_c^k(x)$ as:

\begin{equation}
O_c^k(x) = \frac{\sum_{i,j} \min(S^c_x(i,j), G^k_x(i,j))}{\sum_{i,j} G^k_x(i,j)}.
\end{equation}
where $(i,j)$ are matrix indices. This measures how much of the model's attention for a given class aligns with the regions corresponding to concept $k$. We compute this overlap for all concepts and images in the training set, and for each class $c$, we define the set of important concepts $\mathcal{K}_c$ as those that consistently exhibit high overlap with the saliency maps:

\begin{equation}
\mathcal{K}_c = \left\{ k \ \middle|\ \frac{1}{N_c} \sum_{x \in \mathcal{D}_c} O_c^k(x) > \tau \right\},
\end{equation}
where $N_c$ is the number of training examples in class $c$, $\mathcal{D}_c$ is the set of images in class $c$, and $\tau$ is a threshold that determines the importance of the concept. This procedure functions as a concept discovery module, identifying which local concepts are critical for predicting each class. \\

\subsection{\modelname}
\label{subsec:novelLosses}

As shown in Figure~\ref{fig:pipelineFig}, \modelname~utilizes cross-entropy losses $\mathcal{L}_c$ and $\mathcal{L}_k$ for class and concept labels, respectively, integrated with novel concept saliency alignment and local concept contrastive losses. 

We detail these loss terms below.

\subsubsection{Concept saliency alignment loss}

For each image $x$, we predict important concepts $\mathcal{K}_c$ and enforce alignment between the saliency maps $S_x^k$ for predicted concepts and the GT-maps $G_x^k$. This is encouraged by our proposed concept saliency alignment (\textbf{CSA}) loss:

\begin{equation}
\mathcal{L}_{\text{CSA}} = \frac{1}{|\mathcal{K}_c|} \sum_{k \in \mathcal{K}_c} \| S_x^k - G_x^k \|_2^2.
\end{equation}

Aligning the model's attention with GT-maps enables class-specific reasoning, thereby elucidating the rationale behind predictions by linking them to relevant local features.

\subsubsection{Local concept contrastive loss}

While the CSA loss facilitates explicit localization of concepts, it is equally essential for these concept-level features to exhibit invariance across domain shifts. To achieve domain invariance, we propose a local concept contrastive (\textbf{LCC}) loss, employing a triplet strategy to cluster similar concepts (e.g., \texttt{eyes}) across domains while distinguishing unrelated ones (e.g., \texttt{feathers} and \texttt{ears}).

Let $x$ be an anchor image containing concept $k$, a positive image $x^+$ (an augmentation of $x$ that retains concept $k$), and a negative image $x^-$ (containing a different concept $k'$). For each image, we compute a concept-specific feature vector $f_x^k \in \mathbb{R}^C$, emphasizing the concept’s relevant regions using the GT-map $G_x^k$. Each element $f_x^k(l)$ is computed as:

\begin{equation}
f_x^k(l) = \sum_{i,j} G_x^k \cdot F_x(i,j,l).
\label{eqn:fx_compute}
\end{equation}

where $\cdot$ is an element-wise multiplication operation, $G_x^k$ is a matrix representing the GT-map, $F_x$ represents convolutional feature maps from the last convolution layer, and $(i,j,l)$ denotes the $l^{th}$ channel's $(i,j)$ element. These vectors focus on the concepts of interest (concept $k$ for $x$ and $x^+$, $k'$ for $x^-$), in contrast to the global features used in prior works. The LCC loss is then defined as:

\begin{equation}
\mathcal{L}_{\text{LCC}} = \max(0, d(f_x^k, f_{x^+}^k) - d(f_x^k, f_{x^-}^{k'}) + \alpha),
\end{equation}

where $d(.)$ is the euclidean distance and $\alpha$ is the margin.

\subsection{Test-time Correction}
We establish that our localized, interpretable approach facilitates correction of misclassifications through concept-level feature verification. In this section, we first introduce concept signatures and detail the proposed correction strategy.

\subsubsection{Local concept signatures}
For each concept $k$, we define a concept-signature $p^k \in \mathbb{R}^C$, as its representative vector. We derive $p^k$ by averaging the concept-specific feature vectors $f^k_x$ across all training samples $x \in \mathcal{D}$ containing concept $k$ (denoted as $\mathcal{D}^k$).

\begin{equation}
p^k = \frac{1}{|\mathcal{D}^k|} \sum_{x \in \mathcal{D}^k} f^k_x.
\end{equation}

These vectors act as reference points, helping the model recognize when its attention is aligned with the right concepts during prediction, even when encountering new, unseen domains.

\subsubsection{Detecting and Correcting misclassifications}

\begin{algorithm}[tb]
\scriptsize
    \caption{Iterative Test-time Correction}
    \label{alg:predCorrect}
    \textbf{Input}: Test image $x$, initial class prediction $c^{(0)}$, signatures $p^k$ for each concept $k$\\
    \textbf{Parameter}: Threshold $\delta$, max iterations $T$\\
    \textbf{Output}: Corrected class prediction, $c_{\text{final}}$\\
    \setlength{\baselineskip}{1.5em} 
    \vspace{-10pt}
    \begin{algorithmic}[1]
        \STATE $S_x^{c^{(0)},0} \gets \text{GradCAM}(x, c^{(0)})$  \COMMENT{Initialize saliency map for iteration $t=0$}
        \STATE $x_{\text{masked}}^0 \gets x$
        \FOR{each $k \in  \mathcal{K}_{c^{(0)}}$}
            \STATE $f_x^{k,0} \gets \sum_{i,j} S_x^k \cdot F_x(i,j,:) $ %      \text{ExtractFeatures}(S_x^{k,0})$
            \STATE $d(f_x^{k,0}, p^k) \gets 1 - \frac{f_x^{k,0} \cdot p^k}{\|f_x^{k,0}\| \|p^k\|}$
            \IF{$d(f_x^{k,0}, p^k) > \delta$}
            
           \STATE \textbf{Correction Phase:}
                \FOR{$t \gets 1$ \textbf{to} $T$}
                    \STATE $M^{c^{(t-1)},t} \gets \text{Binarize}(S_x^{c^{(t-1)},t-1})$ %\COMMENT{Binarized mask at iteration $t$}
                    \STATE $x_{\text{masked}}^t \gets x_{\text{masked}}^{t-1} \cdot M^{c^{(t-1)},t}$
                    \STATE $c^{(t)} \gets \text{PredictClass}(x_{\text{masked}}^t)$ %\COMMENT{Predict class at iteration $t$}
                    \STATE $S_{x_{\text{masked}}}^{c^{(t)},t} \gets \text{GradCAM}(x_{\text{masked}}^t, c^{(t)})$
                    \FOR{each $k \in  \mathcal{K}_{c^{(t)}}$}
                        \STATE $f_{x_{\text{masked}}}^{k,t} \gets \sum_{i,j} S_{x_{\text{masked}}}^{k,t} \cdot F_x(i,j,:)  $ %$ \text{ExtractFeatures}(S_{x_{\text{masked}}}^{k,t})$
                        \STATE $d(f_{x_{\text{masked}}}^{k,t}, p^k) \gets 1 - \frac{f_{x_{\text{masked}}}^{k,t} \cdot p^k}{\|f_{x_{\text{masked}}}^{k,t}\| \|p^k\|}$
                        \IF{$d(f_{x_{\text{masked}}}^{k,t}, p^k) \leq \delta$}
                            \STATE \textbf{Return} $c_{\text{final}} \gets c^{(t)}$
                        \ENDIF
                    \ENDFOR
                \ENDFOR
            \ENDIF
            \STATE \textbf{Return} $c_{\text{final}} \gets c^{(0)}$
        \ENDFOR
    \end{algorithmic}
\end{algorithm}

Consider the example of test-time correction in Figure~\ref{fig:pipelineFig}, where the model misclassifies a \emph{person} as a \emph{guitar}. The predicted class \emph{guitar}, involves concepts like \texttt{strings} and \texttt{knobs}, but these are absent in the image. As a result, the model erroneously focuses on irrelevant features (e.g., the person’s legs or the background), as reflected in the GradCAM saliency maps.

To address misclassifications, we employ a two-step approach: possible mistake detection followed by correction. The process is outlined in Algorithm~\ref{alg:predCorrect}. First, the model extracts concept-level saliency maps ($S^k_x$), for all the concepts corresponding to the predicted class, which are then used to compute concept-level features $f^k_x$ (step 4). In Step 5 and 6, we compare the features to the stored concept signatures; a deviation exceeding $\delta$ signals a misclassification.

Once a misalignment is detected, the model enters an iterative refinement phase. The features prominent for the current class level predictions are masked through the corresponding saliency map $S^{c}_x$ (step 9 and 10). The masking process is cumulative, in effect utilizing all masks from the first to the $t^{th}$ iteration. The masked features are then used for subsequent predictions (step 11). This process continues until the concepts corresponding to the predicted class aligns with concept-signatures or the maximum iteration count is reached (steps 8–16 in the algorithm). If alignment is achieved, the predicted class in that iteration is confirmed as the final output.

%% file: sec/4_exps.tex
\section{Experiments}
\label{sec:exps}

\textbf{Datasets:} We conduct experiments on four widely used datasets - PACS~\cite{li2017deeper}, VLCS~\cite{fang2013unbiased}, OfficeHome~\cite{venkateswara2017deep}, and DomainNet~\cite{peng2019moment}, within the DomainBed~\cite{gulrajani2020search} evaluation benchmarks. PACS contains $9\mbox{,}991$ images across $7$ categories in four domains: `sketch’, `photo’, `clipart’, and `painting’. VLCS consists of $10\mbox{,}729$ images over $5$ categories and $4$ domains. The domains are from VOC2007 (V), LabelMe (L), Caltech101 (C), and SUN09 (S), with domain shifts primarily driven by background and viewpoint variations. OfficeHome, with $15\mbox{,}500$ images across $65$ categories, emphasizes indoor classes across `product’, `real’, `clipart’, and `art’ domains.  For DomainNet, we follow prior work~\cite{tan2020class, qi2024generalizing} and use a subset of the $40$ most common classes across `sketch’, `real’, `clipart’, and `painting’. \\

\noindent \textbf{Experimental Setup:}  
We adhere to the SSDG paradigm, training on one \emph{source} domain and testing across three \emph{target} domains, i.e., the model is trained independently four times, with the averaged test accuracies across target domains reported in each case. Training utilizes solely source-domain GT-maps, excluding any target-domain prior knowledge, data or annotations. Minimal augmentations (quantization, blurring, and canny edge), were used to introduce slight perturbations to create the triplets. We use SDv2.1~\cite{rombach2022high} for generating exemplar images and computing cross-attention maps~\cite{agarwal2023star,chefer2023attend}. To ensure a fair comparison, we adopt a ResNet-18 backbone throughout all the experiments. We use the Adam optimizer with an initial learning rate of $1 \times 10^{-4}$ and a warm-up schedule over the first $1000$ steps, after which the rate remains constant. Margin value $\alpha$ is set to $1.0$. The batch size is set to $32$. We empirically set $\delta=0.1$ and cap test-time correction at $T=10$ iterations. \\

\noindent \textbf{Compared Approaches:}
We compare our method with ERM~\cite{gulrajani2020search} baseline and existing approaches that utilize augmentation-based techniques (NJPP~\cite{yuan2022not}, AugMix~\cite{hendrycks2019augmix}, MixStyle~\cite{zhou2021domain}, CutMix~\cite{yun2019cutmix}, RandAugment~\cite{cubuk2020randaugment}), self-supervised and domain adaptation methods (RSC~\cite{huang2020self}, pAdaIn~\cite{nuriel2021permuted}, L2D~\cite{wang2021learning}, RSC+ASR~\cite{fan2021adversarially}), uncertainty modeling (DSU~\cite{li2022uncertainty}, DSU-MAD~\cite{qu2023modality}), attention and meta-learning methods (ACVC~\cite{cugu2022attention}, P-RC~\cite{choi2023progressive}, Meta-Casual~\cite{chen2023meta}), and prompt-based learning (PromptD~\cite{li2024prompt}). The number of methods evaluated differs by dataset.  The discrepancy stems from PACS being the dominant benchmark in prior SSDG works, resulting in a larger number of methods evaluated on it. 

For VLCS, OfficeHome, and DomainNet, we rely on reported results from respective papers, or compute them ourselves (where code was available) to ensure a fair comparison.

\begin{table}[t]
    \centering
     \scriptsize
    \begin{subtable}{1.\linewidth}
        \begin{tabular}{@{}l|cccc|c@{}}
\toprule
Method   Venue & Art   & Cartoon  & Sketch  & Photo  & Average \\ \midrule
ERM   &65.38&64.20& 34.15& 33.65 & 49.35\\
Augmix~\cite{hendrycks2019augmix}~\tabpubb{ICLR'21} &66.54&70.16& 52.48& 38.30 & 57.12\\
RSC~\cite{huang2020self}~\tabpubb{ECCV'20}  &73.40&75.90& 56.20& 41.60 & 61.03\\
Mixstyle~\cite{zhou2021domain}~\tabpubb{ICLR'21} &67.60&70.38& 34.57&37.44 & 52.00\\
pAdaIn~\cite{nuriel2021permuted}~\tabpubb{CVPR'21} &64.96&65.24&32.04&33.66 & 49.72\\
RSC+ASR~\cite{fan2021adversarially}~\tabpubb{CVPR'21} &76.70&79.30& 61.60& 54.60 & 68.30\\
L2D~\cite{wang2021learning}~\tabpubb{ICCV'21} &76.91&77.88&53.66 & 52.29 & 65.93\\
DSU~\cite{li2022uncertainty}~\tabpubb{ICLR'22} &71.54&74.51& 47.75& 42.10 & 58.73\\
ACVC~\cite{cugu2022attention}~\tabpubb{CVPR'22} &73.68&77.39&55.30 & 48.05 & 63.10\\
DSU-MAD~\cite{qu2023modality}~\tabpubb{CVPR'23} &72.41&74.47& 49.60& 44.15 & 60.66\\
P-RC~\cite{choi2023progressive}~\tabpubb{CVPR'23} &76.98&78.54& 62.89& 57.11 & 68.88\\
Meta-Casual~\cite{chen2023meta}~\tabpubb{CVPR'23} &77.13&80.14& 62.55& 59.60 & 69.86\\
ABA~\cite{cheng2023adversarial}~\tabpubb{ICCV'23} &75.69&77.36& 54.12& 59.04 & 66.30\\
PromptD~\cite{li2024prompt}~\tabpubb{CVPR'24}  &78.77&82.69& 62.94& 60.09 & 71.87\\
\midrule
\textbf{\modelname}    & \textbf{86.24}&  \textbf{86.37}      &      \textbf{73.11}   &   \textbf{74.36} & \textbf{80.02} \\ \bottomrule
        \end{tabular}
    \centering
    \caption{PACS}
    \label{tab:pacs_sdg}
    \end{subtable}
     \newline
     \vspace{-0.5em}
     \newline
    \begin{subtable}{1.\linewidth}
    \begin{tabular}{@{}l|cccc|c@{}}
\toprule
Method    & V   & L  & C  & S       & Average \\ \midrule
ERM   &76.72&58.86& 44.95& 57.71 & 59.06\\
Augmix~\cite{hendrycks2019augmix}~\tabpubb{ICLR'19} &75.25&59.52& 45.90& 57.43 & 59.03\\
pAdaIn \cite{nuriel2021permuted}~\tabpubb{CVPR'21}&76.03&65.21&43.17&57.94 & 60.34\\
Mixstyle \cite{zhou2021domain}~\tabpubb{ICLR'21} &75.73&61.29& 44.66&56.57 & 59.06\\
ACVC \cite{cugu2022attention}~\tabpubb{CVPR'22}&76.15&61.23&47.43 & 60.18 & 61.75\\
DSU \cite{li2022uncertainty}~\tabpubb{ICLR'22} &76.93&69.20& 46.54& 58.36 & 62.11\\
DSU-MAD \cite{qu2023modality}~\tabpubb{CVPR'23} &76.99&70.85& 44.78& 62.23 & 63.71\\
\midrule
\textbf{\modelname} & \textbf{82.62}  &  \textbf{86.13} & \textbf{70.12} & \textbf{72.44} & \textbf{77.08}  \\ \bottomrule
    \end{tabular}
    \centering
    \caption{VLCS}
    \label{tab:vlcs_sdg}
    \end{subtable}
     \newline
    \vspace{-0.5em}
     \newline
     \begin{subtable}{1.\linewidth}
\begin{tabular}{@{}l|cccc|c@{}}
\toprule
Method   & Art   & Clipart  & Product  & Real  & Average \\ \midrule
ERM   &57.43&50.83& 48.9& 58.68 & 53.96\\
MixUp~\cite{zhang2017mixup}~\tabpubb{ICLR'18}  &50.41&43.19& 41.24& 51.89 & 46.93\\
CutMix~\cite{yun2019cutmix}~\tabpubb{ICCV'19} &49.17&46.15& 41.2& 53.64 & 47.04\\
Augmix~\cite{hendrycks2019augmix}~\tabpubb{ICLR'19} &56.86&54.12& 52.02& 60.12 & 56.03\\
RandAugment~\cite{cubuk2020randaugment}~\tabpubb{CVPRW'20}&58.07&55.32&52.02&60.82 & 56.56\\
CutOut~\cite{Devries2017ImprovedRO}~\tabpubb{arXiv:1708}&54.36&50.79&47.68&58.24 & 52.77\\
RSC~\cite{huang2020self}~\tabpubb{ECCV'20}&53.51&48.98&47.16&58.3 & 52.73\\
MEADA~\cite{zhao2020maximum}~\tabpubb{NIPS'20} &57.0&53.2&48.81&59.21 & 54.80\\
PixMix~\cite{hendrycks2022pixmix}~\tabpubb{CVPR'22}&53.77&52.68&48.91&58.68 & 53.51\\
L2D~\cite{wang2021learning}~\tabpubb{ICCV'21}&52.79&48.97&47.75&58.31 & 51.71\\
ACVC~\cite{cugu2022attention}~\tabpubb{CVPR'22}&54.3&51.32&47.69 & 56.25 & 52.89\\
NJPP~\cite{yuan2022not}~\tabpubb{ICML'24}  &60.72&54.95& 52.47& 61.26 & 57.85\\
\midrule
\textbf{\modelname}  & \textbf{72.32} &  \textbf{75.13} & \textbf{75.22}  &  \textbf{73.37}  & \textbf{74.01}  \\ \bottomrule
    \end{tabular}
    \centering
    \caption{OfficeHome}
    \label{tab:officehome_sdg}
     \end{subtable}
     \newline
    \vspace{-0.5em}
     \newline
       \begin{subtable}{1.\linewidth}
\begin{tabular}{@{}l|cccc|c@{}}
\toprule
Method   & Clipart   & Painting  & Real  & Sketch  & Average \\ \midrule
ERM & 68.73& 66.12& 68.51 & 69.44&68.2\\
MixUp~\cite{zhang2017mixup}~\tabpubb{ICLR'18} &70.31&64.34& 69.21& 68.82& 68.17\\
CutMix~\cite{yun2019cutmix}~\tabpubb{ICCV'19}&71.52& 63.84& 67.13& 69.41 & 67.98\\
Augmix~\cite{hendrycks2019augmix}~\tabpubb{ICLR'19} &72.37& 62.91& 69.84& 71.22 & 69.09\\
RandAugment~\cite{cubuk2020randaugment}~\tabpubb{CVPRW'20}&69.71& 65.51& 68.36& 66.93 & 67.63\\
CutOut~\cite{Devries2017ImprovedRO}~\tabpubb{arXiv:1708}&70.86& 64.48& 69.92& 71.55 & 69.20\\
RSC~\cite{huang2020self}~\tabpubb{ECCV'20}&68.25& 67.91& 70.76& 66.18 & 68.28\\
PixMix~\cite{hendrycks2022pixmix}~\tabpubb{CVPR'22}&72.12& 63.51& 71.34& 65.46 & 68.12\\
NJPP~\cite{yuan2022not}~\tabpubb{ICML'24}  &76.14&69.24& 76.61& 71.21 & 73.3\\
\midrule
\textbf{\modelname} & \textbf{82.42}&\textbf{80.37}&\textbf{84.15}&\textbf{81.61}      &  \textbf{82.14} \\ \bottomrule
    \end{tabular}
    \centering
    \caption{DomainNet}
    \label{tab:domainNet_sdg}
     \end{subtable}
    \caption{ SSDG classification accuracies (\%) on PACS, VLCS, OfficeHome and DomainNet datasets, with ResNet-18 as backbone. Each column title indicates the source domain, and the numerical values represent the average performance in the target domains.}
    \label{tab:main_results}
\end{table}

\subsection{Main Results}
Table~\ref{tab:main_results} provide a comparison of average classification accuracy for our approach against existing methods across the PACS, VLCS, OfficeHome, and DomainNet datasets. Each column in these tables represents a source domain used for training, with the numerical values indicating the average accuracy on the three target domains. The last column presents the average of the four columns. 

\modelname~decisively outperforms existing approaches across all datasets. It achieves average accuracy gains of $8.33$\%, $13.37$\%, $16.16$\%, and $8.84$\% over the second-best approach on PACS, VLCS, OfficeHome, and DomainNet, respectively. Table~\ref{tab:vlcs_sdg} details the performance on VLCS, where domain shifts primarily stem from background and viewpoint variations, with scenes spanning urban to rural and favoring non-standard viewpoints. Data augmentation based methods, which focus on style variation, yield limited gains on VLCS relative to PACS. Nonetheless, \modelname~secures substantial improvements, notably achieving a $25.34$\% gain over DSU-MAD when Caltech101 (C) serves as the source domain. Another noteworthy observation is that \modelname~maintains strong performance across varying class counts, from VLCS's $5$ to OfficeHome's $65$, underscoring the scalability of the proposed approach.

\begin{table}[t]
\scriptsize
\resizebox{0.47\textwidth}{!}{%
\begin{tabular}{@{}l|cccc|c@{}}
\toprule
Method & Art & Cartoon & Sketch  & Photo & Average \\ \midrule
$\mathcal{L}_c$   &65.38&64.20& 34.15& 33.65 & 49.35\\
$\mathcal{L}_c+\mathcal{L}_k$   & 65.47&64.12&35.44&33.81&49.71\\
$\mathcal{L}_c+\mathcal{L}_k+\mathcal{L}_{CSA}$   & 66.11& 64.23&35.14&34.27&49.93\\
$\mathcal{L}_c+\mathcal{L}_k+\mathcal{L}_{CSA}+\mathcal{L}_{LCC}$   &80.28 &82.91&66.12& 65.37& 73.67\\ \midrule

+ test-time correction & \textbf{86.24}&  \textbf{86.37}      &      \textbf{73.11}   &   \textbf{74.36} & \textbf{80.02}

\\ \bottomrule
\end{tabular}%
}
\vspace{-8pt}
    \caption{Ablation study (\%) on PACS.}
    \vspace{-4pt}
    \label{tab:pacs_ablation}
\end{table}

\subsection{Ablations}

{\bf Components of TIDE:} Ablation experiments on the PACS dataset are conducted to assess the contribution of each loss term on the overall performance. In Table~\ref{tab:pacs_ablation}, we present classification accuracy by incrementally adding components to the training pipeline, demonstrating the contribution of each element to the model's performance. We observe that the introduction of the concept classification loss $\mathcal{L}_k$ and concept saliency alignment loss $\mathcal{L}_{CSA}$ does not significantly affect the classifier's accuracy on the test domain. However, these components are integral to our approach, enabling test-time correction and, in turn, enhancing both performance and model interpretability. We observe that the introduction of the $\mathcal{L}_{LCC}$ loss leads to a substantial increase in test domain accuracy, clearly demonstrating its effectiveness in fostering domain invariance. Finally, we compare results before and after test-time correction, highlighting that even prior to correction, our method achieves state-of-the-art performance across all source domains on PACS, underscoring the strength of our approach.

\begin{figure}[t]
    \centering
    \includegraphics[width = \linewidth]{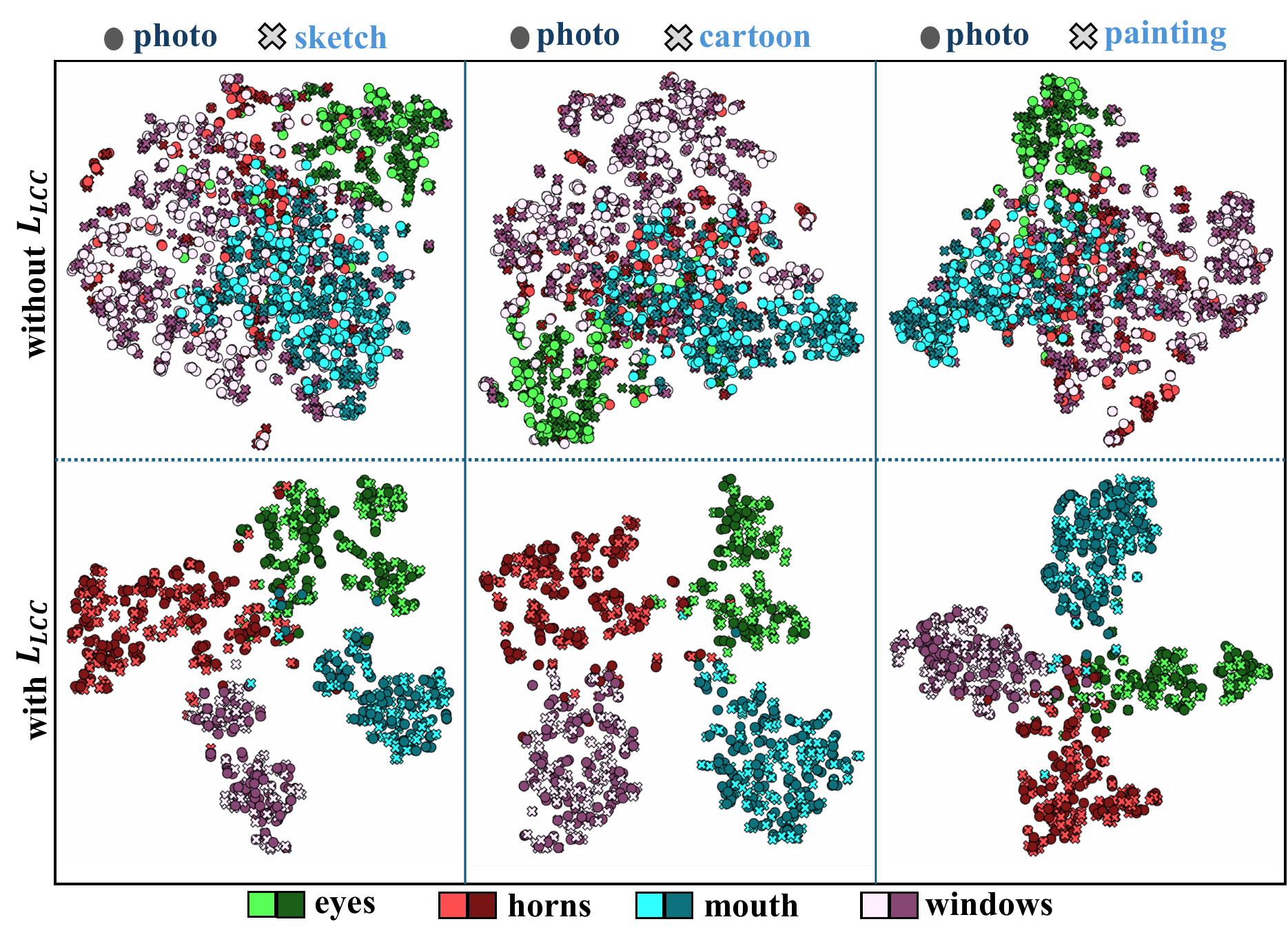}

    \caption{t-SNE visualizations to demonstrate impact of $\mathcal{L}_{LCC}$. Each column represents a test domain (Sketch, Cartoon, Painting), with the top row showing t-SNE plots without $\mathcal{L}_{LCC}$ applied and the bottom one with it. Please zoom in for optimal viewing.}

    \label{fig:tsneVis}
\end{figure}

\begin{figure*}[t]
    \centering
    \includegraphics[width = 0.85\linewidth]{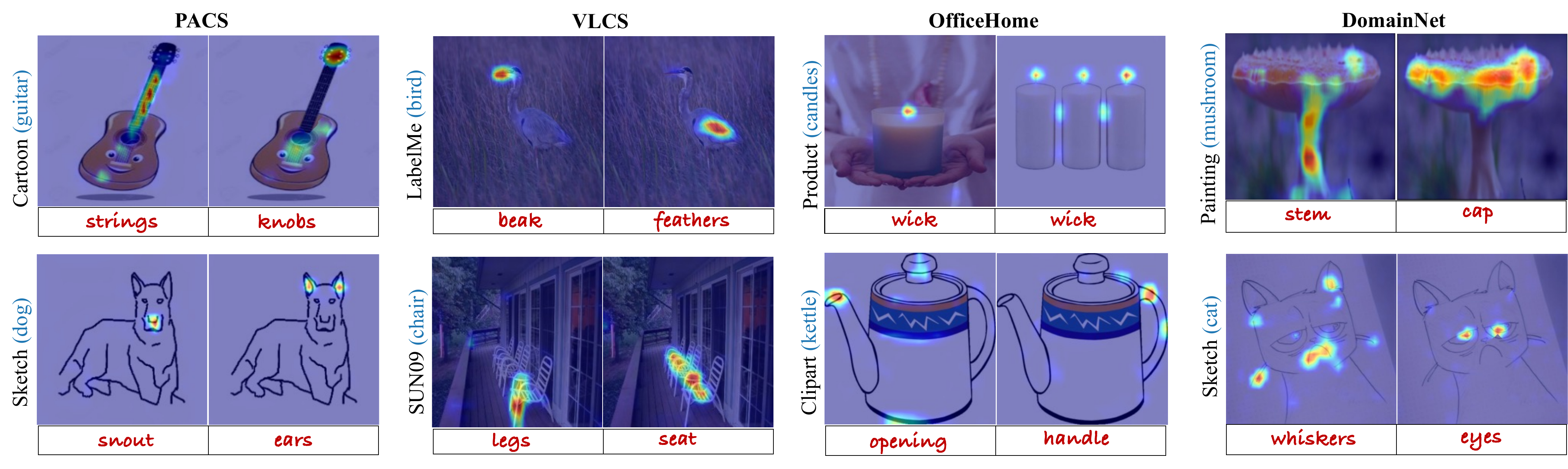}
    \caption{Illustrative examples of concept level GradCAM maps corresponding to \modelname's predictions, across the four studied datasets. The concept names are displayed beneath the maps, with the target domain and predicted class indicated on the left. More results in supplement.}
    \label{fig:qualLocalMain}
\end{figure*}

\begin{figure}[t]
    \centering
    \includegraphics[width = 0.83\linewidth]{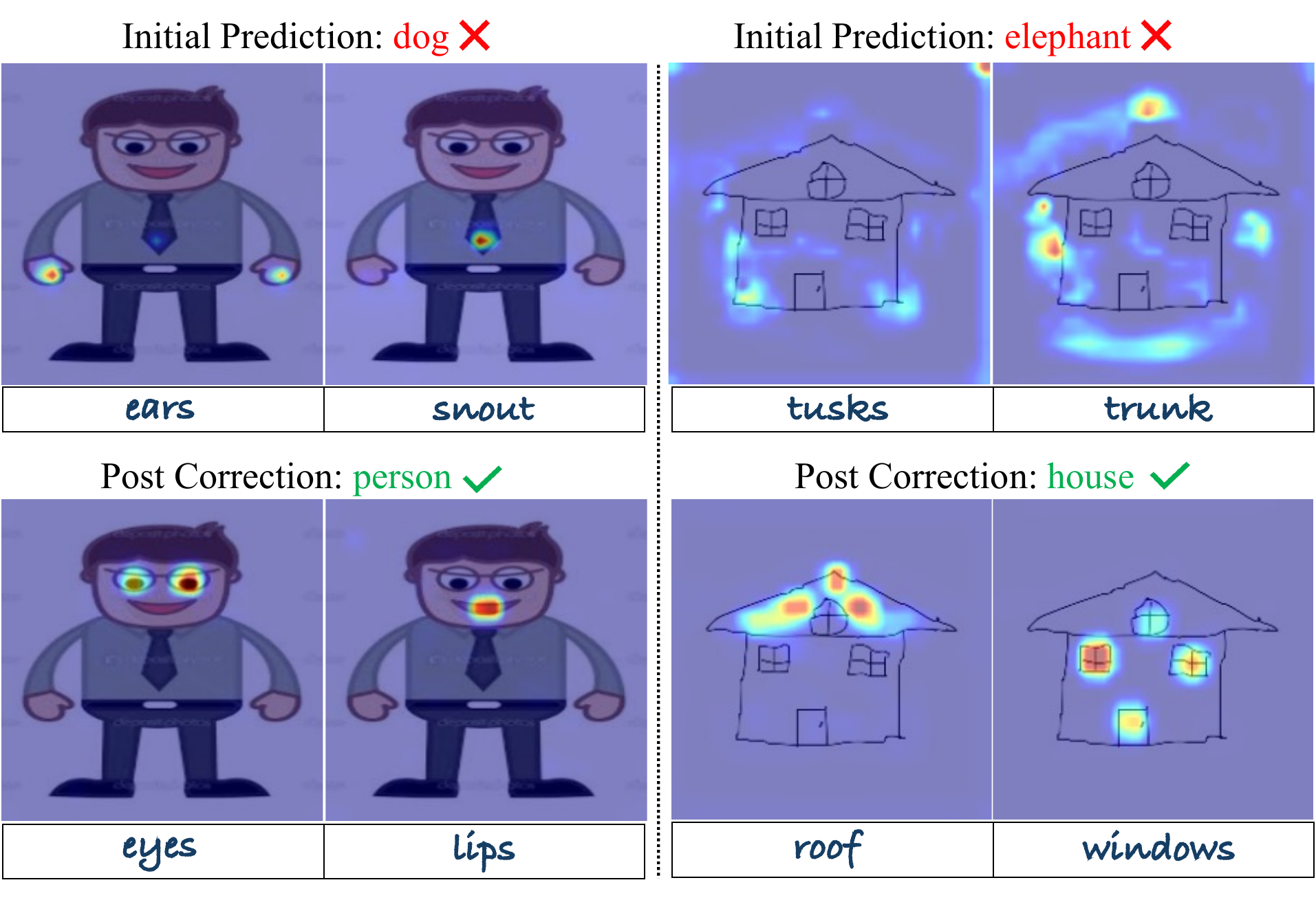}
    \caption{The top row shows initial class predictions and GradCAM maps for the concepts, while the bottom row presents the results after test-time correction.}
    \label{fig:qualCorrectionMain}
\end{figure}

{\bf Error analysis of test-time correction:} For this analysis, we train TIDE on the photo domain of PACS and test it on the sketch domain. The model gives an initial accuracy of $74.79$\%, which improves to $82.29$\% post test-time correction. In $72.2$\% of test samples, \modelname~does not invoke test-time correction, with class predictions correct in $93.8$\% of these cases, demonstrating substantial reliability. The supplement provides examples of cases where the model misclassifications are not picked-up in the signature matching step of test-time correction. In the remaining $27.8$\% of cases where correction is initiated, $52.5$\% successfully converge to the correct classification, significantly contributing to the final post-correction accuracy of $82.29$\%.

{\bf t-SNE plots: } To demonstrate the impact of $\mathcal{L}_{LCC}$, we present t-SNE visualizations in Figure~\ref{fig:tsneVis}, showing the distribution of concept-specific vectors (as computed in Equation~\ref{eqn:fx_compute}). We individually plot the source domain (photo) along with sketch, cartoon, and painting domains as targets. Each case includes two plots: one without $\mathcal{L}_{LCC}$ (top row) and one with it (bottom row). We observe that with $\mathcal{L}_{LCC}$, concept samples (e.g., \texttt{mouth}) align closely across domains, while distinct separation occurs between different concepts (e.g., \texttt{mouth} and \texttt{horns}). Without $\mathcal{L}_{LCC}$, cluster separation and alignment across domains are weaker, highlighting its role in improving intra-concept compactness and inter-concept separability.

\subsection{Qualitative Results}
{\bf Concept Localization:} We train the SSDG model on photo domain for PACS, OfficeHome, and DomainNet and on Caltech101 for VLCS. The predicted class, concepts and corresponding saliency maps are shown in Figure~\ref{fig:qualLocalMain}. For each target class (e.g., mushroom, kettle, guitar), the model reliably highlights key concept-specific regions (e.g., \texttt{stem}, \texttt{handle}, \texttt{strings}) essential for classification.  The model effectively isolates key features across diverse contexts, such as the \texttt{legs} and \texttt{seat} of a chair in complex zoomed-out scenes, the \texttt{beak} and \texttt{feathers} of camouflaged birds, the \texttt{eyes} and \texttt{whiskers} in deformed sketches of cats, and varying feature sizes, with the \texttt{wick} occupying a small area and the \texttt{mushroom cap} spanning a larger one.

\noindent{\bf Test-time Correction:} The Figure~\ref{fig:qualCorrectionMain} visually illustrates the efficacy of \modelname's test time correction abilities, by comparing the initial and corrected results. The model initially misclassifies the images as dog and elephant, resulting in poorly aligned concept-level saliency maps for corresponding concepts i.e. \texttt{ears}, \texttt{snout}, \texttt{tusks} and \texttt{trunk}. \modelname~detects and rectifies such errors during inference, leading to accurate classification accompanied by precise concept localization. This qualitative evaluation reinforces the robustness of our approach in refining predictions and generating reliable, class-specific concept maps.

%% file: sec/5_conclusion.tex
\section{Conclusion}

In this work, we considered the problem of single-source domain generalization and observed that the current state-of-the-art methods fail in cases of semantic domain shifts primarily due to the global nature of their learned features. To alleviate this issue, we proposed TIDE, a new approach that not only learns local concept representations but also produces localization maps for these concepts. With these maps, we showed we could visually interpret model decisions while also enabling correction of these decisions at test time using our iterative attention refinement strategy. Extensive experimentation on standard benchmarks demonstrated substantial and consistent performance gains over the current state-of-the-art. Future work will explore methods to generate confidence scores grounded in TIDE's concept verification strategy and would explore application of TIDE on other (e.g. fine-grained) classification datasets.

%% file: sec/supplementary.tex
\clearpage
\setcounter{page}{1}

\begin{appendices}
    
\section{}
\label{sec:appendix}

In Section~\ref{subsec:testError}, we show examples where the proposed signature verification step does not detect incorrectly predicted classes. In Section~\ref{subsec:njpp}, we qualitatively compare against an additional baseline to highlight the limitations of only learning global class level features, while showing that our method effectively captures local concepts. In Section~\ref{subsec:qualLocal}, we show additional results for concept localization to show further evidence for how our model effectively captures key features across diverse contexts and styles. In Section~\ref{subsec:diftSupp}, we show additional samples of images annotated via our proposed pipeline across all domains from all datasets namely PACS~\cite{li2017deeper}, VLCS~\cite{fang2013unbiased}, OfficeHome~\cite{venkateswara2017deep}, and DomainNet~\cite{peng2019moment}. Finally, we provide details of prompting GPT3.5~\cite{brown2020language} in Section~\ref{subsec:llmDetails}.

\subsection{Signature Verification Errors}
\label{subsec:testError}
\begin{figure}[h]
    \centering
    \includegraphics[width=\linewidth]{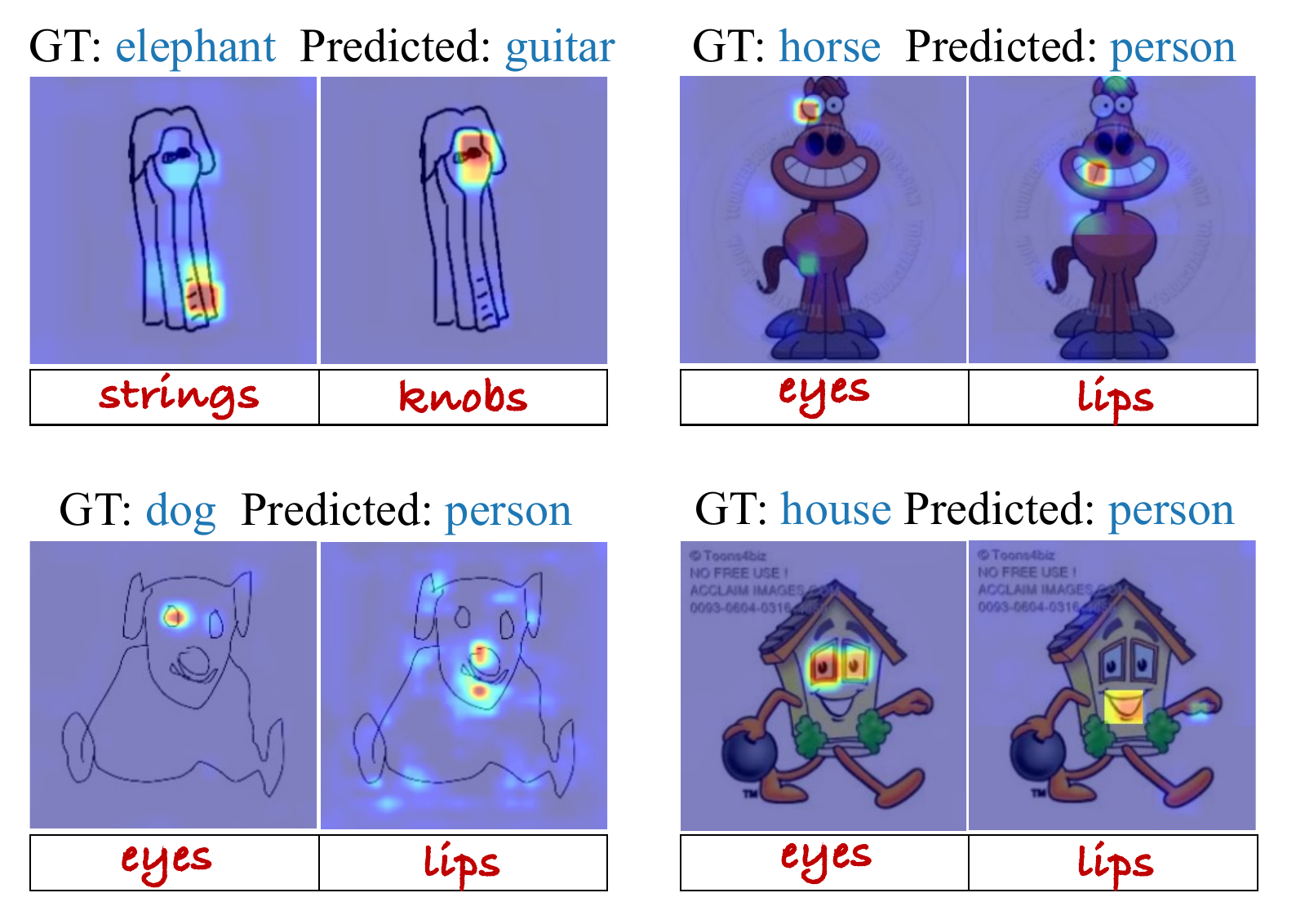}

 \caption{Signature Verification Errors.} 

    \label{fig:corrnErrors}
\end{figure}

In this section, we provide examples where the model's misclassifications are not detected during the test-time signature matching step. We demonstrate that this is prevalent particularly when regions in the input image resemble concepts from other classes. For example, as shown in Figure~\ref{fig:corrnErrors}, the first row/first column illustrates an image of an elephant from the sketch domain of the PACS dataset. The model incorrectly classifies the image as a guitar due to certain visual similarities: the elephant’s \texttt{trunk}, with its line-like patterns, is interpreted as the \texttt{strings} of a guitar, and the unusual, dot-like appearance of the eyes is misidentified as guitar \texttt{knobs}. Similarly, additional examples illustrate this phenomenon: a dog, horse, and house are all misclassified as a person across various domains due to the presence of person-like concepts such as \texttt{eyes} and \texttt{lips}. Notably, the house from the cartoon domain features drawn elements resembling \texttt{eyes} and \texttt{lips}, which the model identifies as human facial features. Since these highlighted regions correspond to semantically plausible features for the predicted classes (guitar/person), the detection step does not flag this as an error, mistakenly believing the prediction to be correct.

\subsection{Additional Baseline}
\label{subsec:njpp}
\begin{figure}[t]
    \centering
    \includegraphics[width=\linewidth]{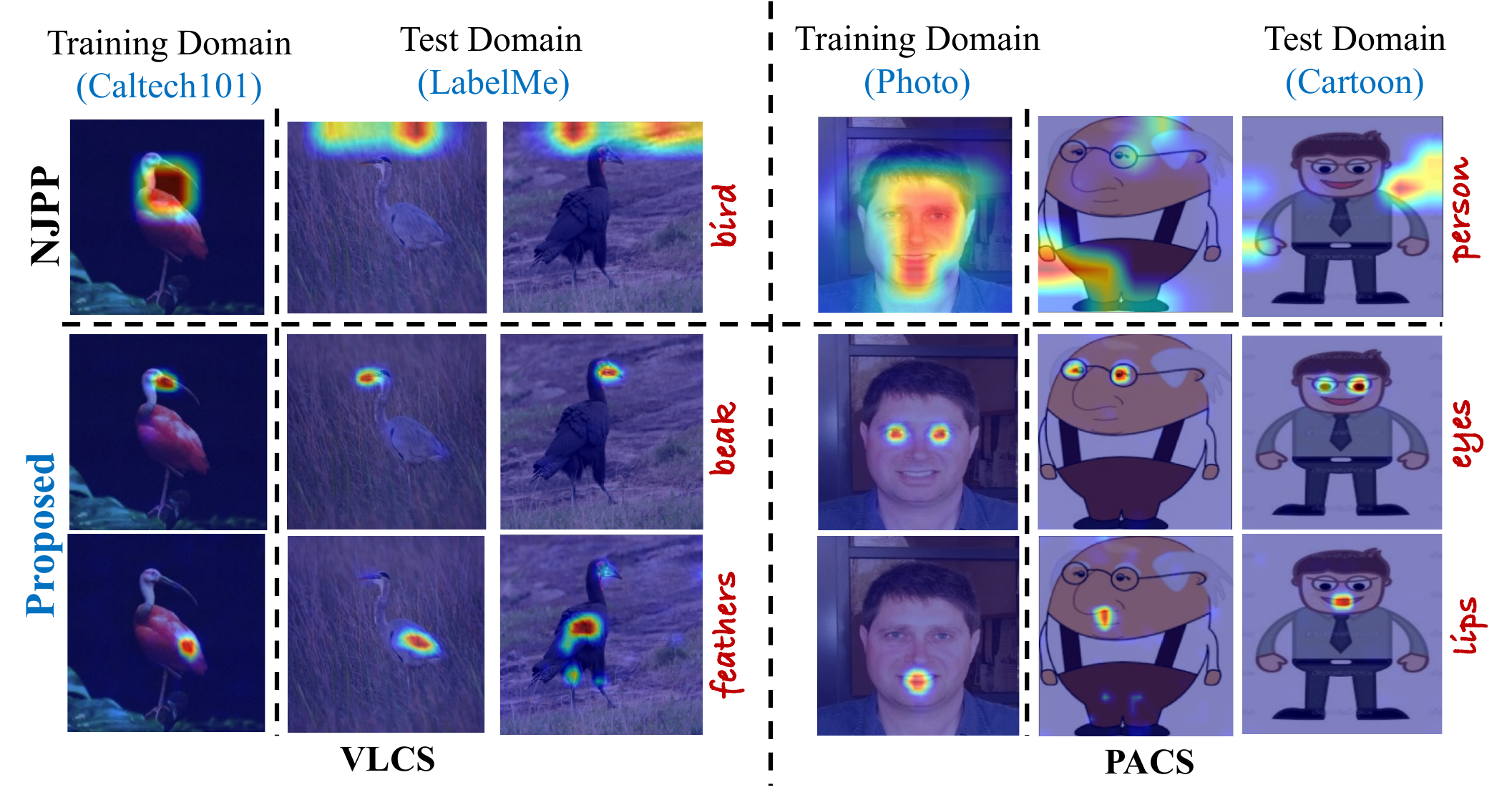}

 \caption{Samples from VLCS (left) and PACS dataset (right) across domain shifts, corresponding to bird and person classes. The first row shows GradCAM~\cite{selvaraju2017grad} saliency maps computed using the baseline NJPP~\cite{yuan2022not}, while the second and third row refers to the proposed method where the model highlights key concepts across domains consistently.} 

    \label{fig:njppSupp}
\end{figure}

We show comparisons with an additional baseline NJPP~\cite{yuan2022not} in Figure~\ref{fig:njppSupp}. One can note that similar to the baseline ABA~\cite{cheng2023adversarial} demonstrated in Figure $1$ in the main paper, NJPP too struggles to maintain consistent attention under domain shifts, often focusing on irrelevant regions (see first row). On the other hand, our method in the second and third rows accurately highlights stable local features like \texttt{beak} and \texttt{feathers}. 

\subsection{Additional Concept Localization Results}
\label{subsec:qualLocal}
\begin{figure*}[t!]
    \centering
    \includegraphics[width=\linewidth]{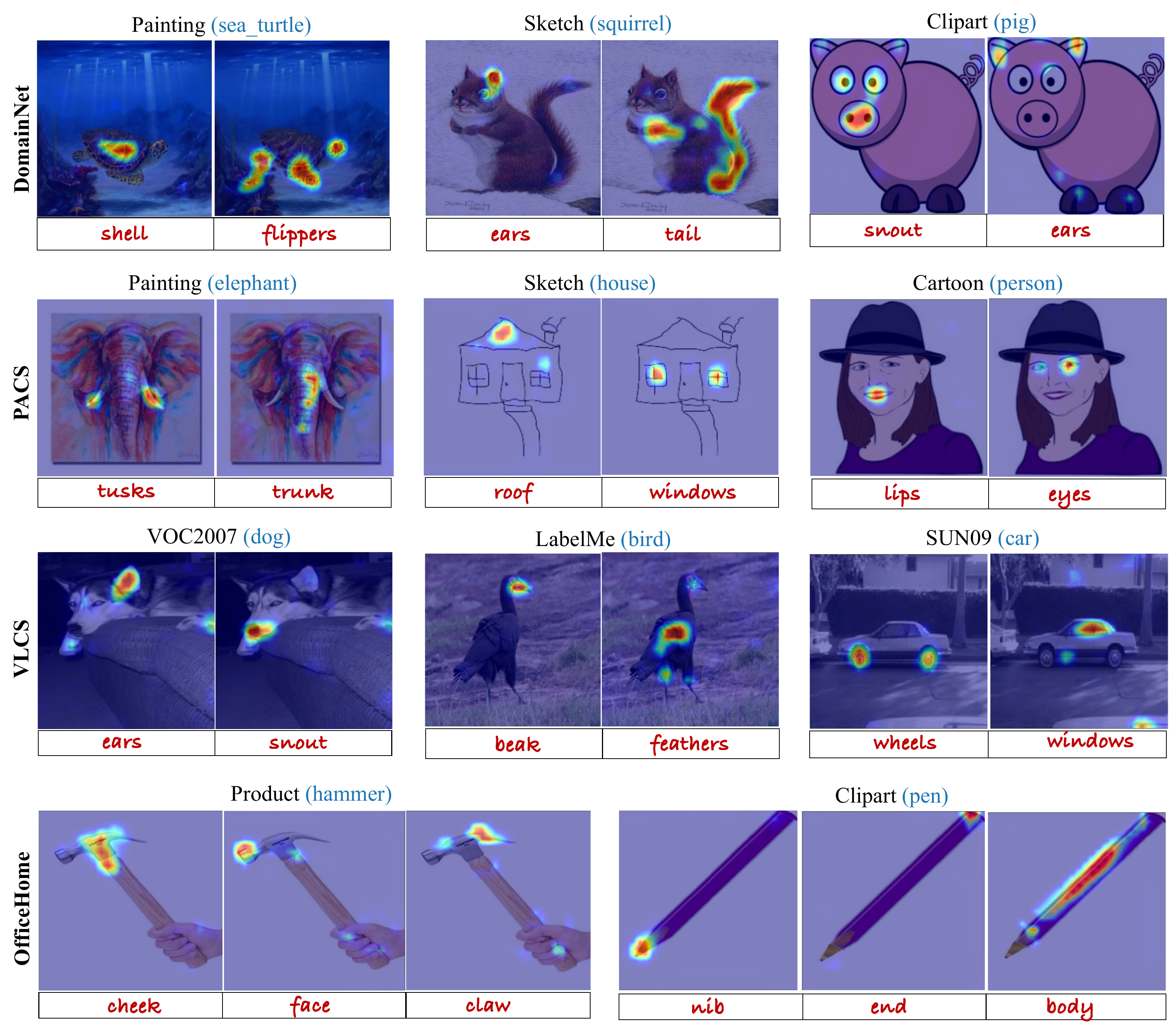}

 \caption{Additional Qualitative Results for Concept Localization.} 

    \label{fig:qualSupp}
\end{figure*}

We show additional results showcasing the model’s ability to localize key concept-specific regions across diverse datasets and domains. For instance, consider the examples in Figure~\ref{fig:qualSupp}, where the model highlights the \texttt{shell} and \texttt{flippers} of sea turtle in paintings (DomainNet), the \texttt{windows} and \texttt{roof} of house in sketch domain (PACS), and the \texttt{wheels} and \texttt{window} of car in a cluttered background (VLCS). Additionally, it maintains consistent focus across varying visual styles, such as abstract representations in cartoons or pencil strokes in sketches. 

\subsection{Additional Annotation Results}
\label{subsec:diftSupp}
\begin{figure*}[t!]
    \centering
    \includegraphics[width=\linewidth]{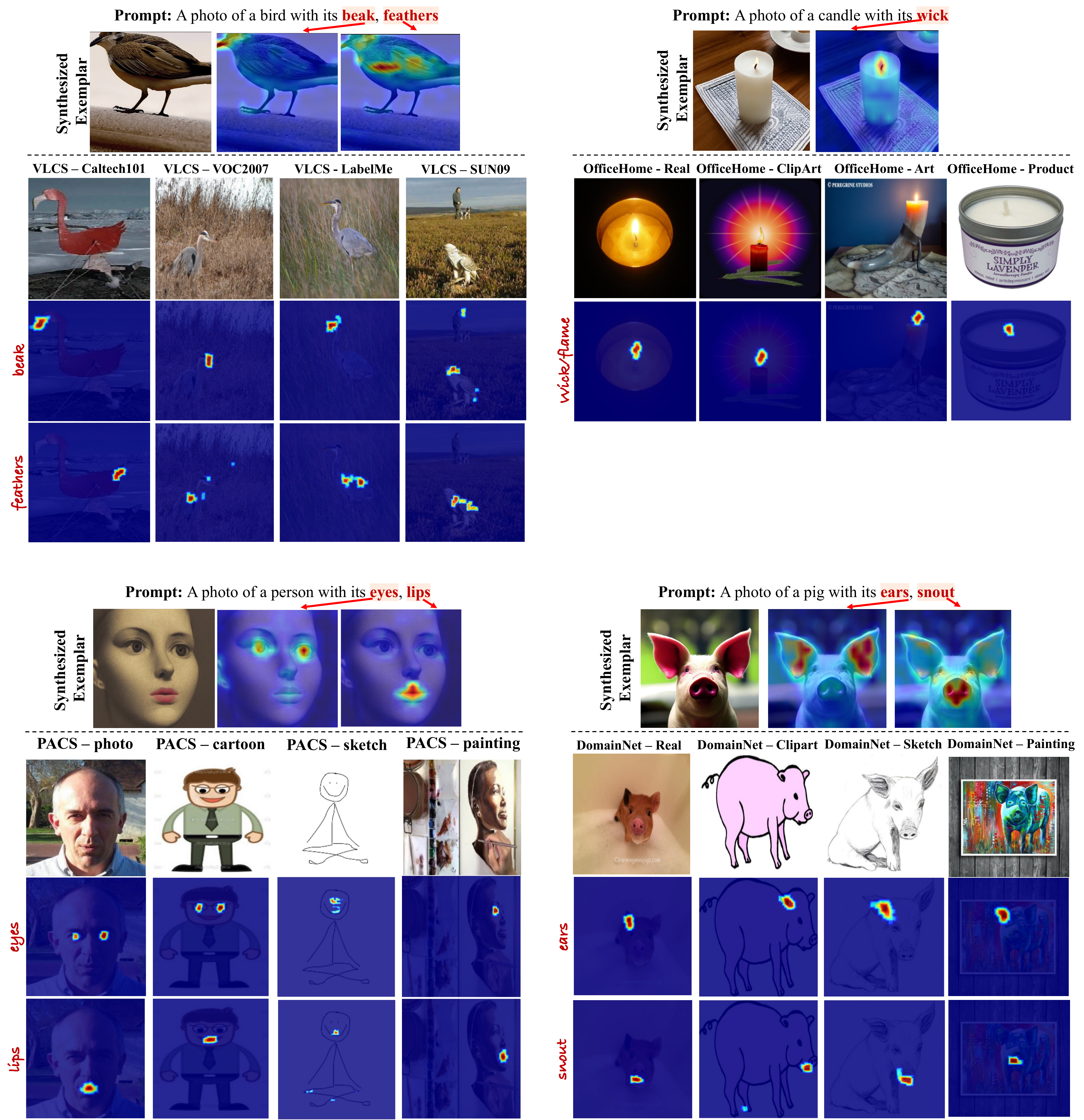}

 \caption{For each dataset (PACS, VLCS, OfficeHome, and DomainNet), we show an example of concept saliency maps (for \texttt{ear} and \texttt{mouth}) transferred from a single synthesized exemplar image to various target domain images. The figure includes one example per dataset, illustrating how the concept saliency maps align across different domains using diffusion features.} 

    \label{fig:diftSupp}
\end{figure*}

We show additional concept-level annotations generated across various datasets and domains in Figure~\ref{fig:diftSupp} using the pipeline proposed in Section $3.1$ in the main paper.

\subsection{LLM Prompting Details}
\label{subsec:llmDetails}
As mentioned in Section $3.1$ in the main paper, we used GPT-3.5~\cite{brown2020language} to generate a list of distinctive, stable features that are semantically relevant for classification. We instruct the LLM to focus on features that are not specific to any particular domain. This means it avoids features that might appear in some domains but not in others. For example, \texttt{fur} may be present in photos of cats but is less likely to appear in sketches of cats. Below, we provide the exact prompt used:\\

\begin{figure}[h]
\centering
\noindent\begin{minipage}{0.5\textwidth}
\mdfsetup{%
middlelinewidth=1pt,
backgroundcolor=cyan!10,
innerleftmargin=0.5cm,
innerrightmargin=0.5cm,
roundcorner=15pt}
\begin{mdframed}
\vspace{0.02em}
 
\textbf{Prompt:} List the most visually distinctive and static features of a \texttt{classname} that a classification model would rely on for accurate identification. Focus only on domain-agnostic features that are intrinsic to the object itself and truly discriminative for the class, avoiding any features that may be related to the environment or context in which the object is typically found. 
 
\end{mdframed}
\end{minipage}
\label{fig:combcap_prompt}
\vspace{0.02em}
\end{figure}

\textbf{Example Outputs: } 
\begin{itemize}
    \item \textit{cat}: \texttt{whiskers}, \texttt{eyes}, \texttt{ears}
    \item \textit{dog}: \texttt{snout}, \texttt{ears}, \texttt{tail}
    \item \textit{bird}: \texttt{beak}, \texttt{feet}, \texttt{feathers}
    \item \textit{squirrel}: \texttt{tail}, \texttt{ears}, \texttt{claws}
    \item \textit{hammer}: \texttt{claw}, \texttt{cheek}, \texttt{face}   
\end{itemize}

Once we identified the key concepts for each class as above, we formulated a structured prompt to guide the diffusion model in generating synthetic images that highlight these concepts e.g. the synthesized exemplar image in Figure $3$ and $4$ in the main paper. The template for the prompt is as follows:\\
\textbf{Template Text:} \textit{Generate a photo of a \texttt{classname} with its \texttt{concept 1}, \texttt{concept 2}, ..., and \texttt{concept n}.}\\
\textbf{Example Prompts:} 
\begin{itemize}
\item For the class \textbf{dog}:
\textit{Generate a photo of a \texttt{dog} with its \texttt{snout}, \texttt{ears} and \texttt{tail}.}
\item For the class \textbf{chair}:
\textit{Generate a photo of a \texttt{chair} with its \texttt{seat}, \texttt{legs}, and \texttt{backrest}.}
\end{itemize}

\end{appendices}